\documentclass[letterpaper, 10 pt, conference]{ieeeconf}  

\IEEEoverridecommandlockouts                              

\usepackage{color}
\usepackage{url}
\usepackage[T1]{fontenc}

\title{\LARGE \bf
RSG-Net: Towards Rich Sematic Relationship Prediction for Intelligent Vehicle in Complex Environments
}
\author{Yafu Tian$^{1,4}$, Alexander Carballo$^{2,3}$, Ruifeng Li$^{4}$ and Kazuya Takeda$^{1,2,3}$
\thanks{$^{1}$ Graduate School of Informatics, Nagoya University, Furo-cho, Chikusa-ku, Nagoya 464-8603, Japan.}%
\thanks{$^{2}$ Institute of Innovation for Future Society, Nagoya University, Furo-cho, Chikusa-ku, Nagoya 464-8601, Japan.}%
\thanks{$^{3}$ Tier IV Inc., Nagoya University Open Innovation Center, 1-3, Mei-eki 1-chome, Nakamura-Ward, Nagoya, 450-6610, Japan.}%
\thanks{$^{4}$ State Key Laboratory of Robotic and Intelligent System, Harbin Institute of Technology, Heilongjiang, China. \hfill\break%
Email:{\tt\scriptsize\underline{yafu.tian@g.sp.m.is.nagoya-u.ac.jp}} (corresponding author), \tt\scriptsize{alexander@g.sp.m.is.nagoya-u.ac.jp, lrf100@hit.edu.cn, 
kazuya.takeda@nagoya-u.jp.}}}%

\usepackage{comment}
\usepackage{flushend} 

\usepackage{xcolor} 
\usepackage{color,soul}
\usepackage{ulem}
\usepackage{adjustbox}
\newif\ifversionwithcorrections      \versionwithcorrectionstrue %
\ifversionwithcorrections

\newcommand{\removedtext}[1]{\adjustbox{bgcolor=red!30,varwidth={\linewidth}}{\st{#1}}}
\newcommand{\rremovedtext}[1]{\adjustbox{bgcolor=red!30,varwidth={\linewidth}}{\hsout{#1}}}

\else

\newcommand{\removedtext}[1]{}
\newcommand{\rremovedtext}[1]{}

\fi

\usepackage{amsmath}
\DeclareMathOperator*{\argmax}{arg\,max}

\begin{document}

\maketitle
\thispagestyle{empty}
\pagestyle{empty}

\begin{abstract}

    Behavioral and semantic relationships play a vital role on intelligent self-driving vehicles and ADAS systems. 
    Different from other research focused on trajectory, position, and bounding boxes, relationship data provides a human understandable description of the object's behavior, and it could describe an object's past and future status in an amazingly brief way. Therefore it is a fundamental method for tasks such as risk detection, environment understanding, and decision making. 
    In this paper, we propose RSG-Net (Road Scene Graph Net): a graph convolutional network designed to predict potential semantic relationships from object proposals, and produces a graph-structured result, called ``Road Scene Graph''. The experimental results indicate that this network, trained on Road Scene Graph dataset, could efficiently predict potential semantic relationships among objects around the ego-vehicle. 
\end{abstract}

\section{INTRODUCTION}

Although autonomous driving systems have advanced significantly in recent years, one open problem is: how human drivers finish their task perfectly without high accuracy velocity/distance/position sensors? Although a human driver only provides very rough geometry result, it overwhelms autonomous driving systems on multiple tasks such as risk detection, scene understanding, vehicle behavior prediction, lane changing, etc. For example, most human drivers can predict a potential car accident in front of them and avoid it, while that is quite a hard task for AI. 

An answer for this question is, human drivers can infer the relationship among surrounding objects with enormous accuracy. In contrast, so far the application of road semantic information is too simple to predict any semantic relationships. And in a complex environment like crossroads and parking lots, these relationships is even more important than the high-accuracy geometry, as these relationships indicate the previous and future status of objects. If such relationships could be captured by a model, the safety and comfort of autonomous system could be increased significantly. And, if failure happens, that information could be helpful for finding out its purpose. 

In recent years, there is a great amount of research focusing on behavior detection and prediction. However, most of them are focused on trajectory prediction for pedestrians or vehicles. Only few research works mentioned such relational data. One reason for this is that trajectory prediction can directly increase the performance of self-driving tasks, while relational data, which are highly unstructured, are hard to process by state-of-art deep-learning models. Also, there are only a few datasets, like Honda Research Institute Driving Dataset (HDD)\cite{ramanishka2018CVPR} and Road Scene Graph Dataset \cite{tian2020road}, which was based on nuScenes dataset \cite{caesar2020nuscenes}, to include relational data.

\begin{figure}[htbp]
    \centering
    \includegraphics[width=1\linewidth]{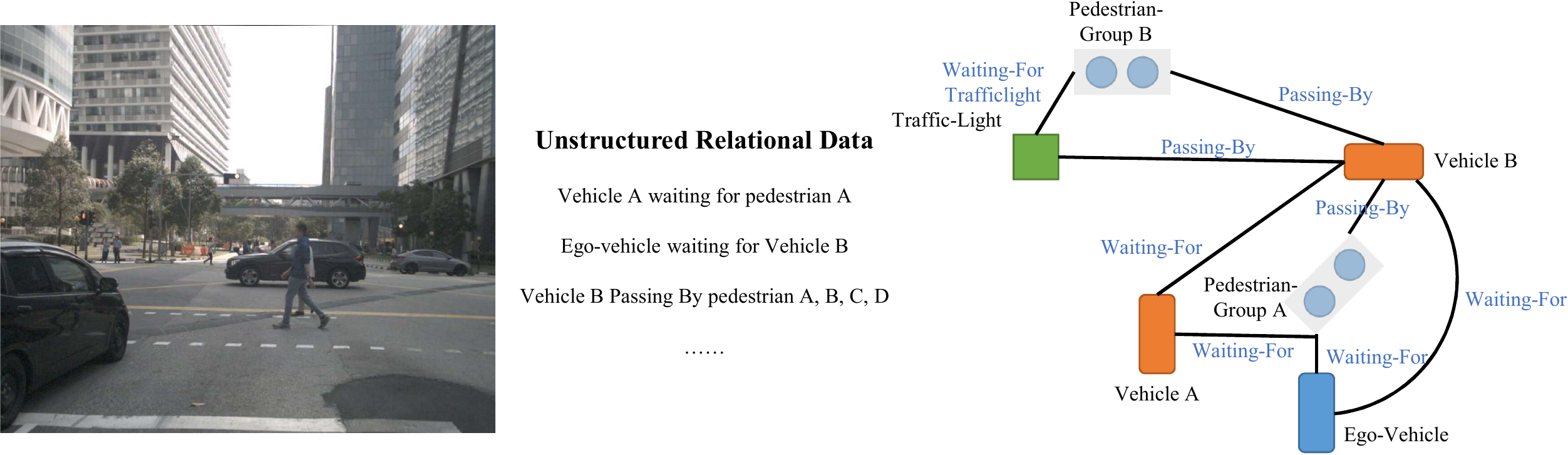}  
    \caption{  Semantic relationships in an image recorded by vehicle recorder. Nodes in the right graph stand for the objects. And edges for pairwise relationships. Besides, we use the concept of ``group'' to describe multiple objects which have similar behavior.}
    \label{fig-intro}
\end{figure}

In this paper, we propose RSG-Net (Road Scene Graph Net), graph convolutional network (GCN), designed to predict relationship information from surrounding objects' proposals. Here, ``objects'' refer to three main categories: vehicle, pedestrian, and obstacle. This model takes those objects' bird-view bounding box as input and proposes corresponding road scene graph, which is a topological graph in which nodes refer to objects and edges refer to the type of semantic relationship between objects. Fig. \ref{fig-intro} illustrates a sample of rich semantic relationships provided by our model, and these relationships play a vital role in driver's decision making. Furthermore, as many objects, like pedestrians or vehicles nearby, tend to move in the same mode, thus forming a group or cluster, their speed, direction, and behavior stay relatively the same. We brought in the ``group'' concept to keep the result simple, and user-friendly.

In summary, our key contribution are as follows:

\begin{itemize}
    \item  We propose RSG-Net (Road Scene Graph Net) to predict potential semantic relationship from object proposals and outputs road scene graph for relationship prediction.
    \item  Performance evaluations of RSG-Net using our Road Scene Graph Dataset\cite{tian2020road}.
    \item  A benchmarking model of our Road Scene Graph Dataset to generate graphs from scratch.
\end{itemize}

This paper is structured as follows: Section~\ref{s:relatedwork} presents state-of-the-art works on drive scene understanding, relationship prediction and related datasets. Section~ \ref{s1:methods} describes in detail our Road Scene Graph Net method, the model and evaluation metrics, while Section~\ref{s:experiment} presents the results of different evaluation experiments. Finally, this paper is concluded in Section~\ref{s:conclusions}.

\section{RELATED WORK}
\label{s:relatedwork}
\subsection{Driving Scene Understanding: Tasks and methods}

Significant progress has been made in many self-driving tasks like object detection, semantic segmentation and 3D pose estimation \cite{kuo2015deepbox, cho2014multi, xu2018pointfusion}. However, when it comes to semantic areas like behavioral prediction, relationship prediction, or potential risk evaluation, these approaches performance cannot catch up with human drivers yet. Recently, several papers have applied semantic information on intelligent vehicles, such as vehicle/pedestrian behavior prediction, trajectory prediction \cite{liang2020learning}, potential risk detection. Besides that, researches are focusing on explaining the intention of human driver or self-driving system \cite{kim2017interpretable, kim2019grounding}, and then generate natural language description for it\cite{ kim2018textual}. 

Since late 2020, some researchers are focusing on graph structure scene representation in the self-driving area. In \cite{li2020learning}, they proposed an interaction graph, which means two objects are connected if there might be an ``interaction'' between them. And it has been proved that this graph model, although quite simple, can increase the performance of driver behavior annotation. In \cite{mylavarapu2020understanding}, they generate another kind of graph representation, where edges stands for the relative motion between two objects. And in \cite{yu2020scene}, the graph is more than a plural graph which connects object with lanes. In our own previous work \cite{tian2020road}, we proposed road scene graph concept, and proposed a model that takes this graph as input and predict the graph in the next few seconds.

Undoubtedly, a general problem of these task-specific semantic researches is that datasets are insufficient. Some data like driving behavior, risky scene, and even car accidents, are extremely hard to obtain. Luckily, in 2019 and 2020, an increasing amount of high-quality semantic driving dataset appeared. Some of them are common datasets, like nuScenes \cite{caesar2020nuscenes}, sematic KITTI \cite{behley2019semantickitti} and Waymo \cite{sun2020scalability}. And others, although with less amount of data, are task-specific datasets, e.g. Road Hazard Stimuli dataset \cite{wolfe2020rapid} for road hazard and risk annotation, Honda HDD dataset \cite{ramanishka2018CVPR} for behavior prediction, and our Road Scene Graph Dataset \cite{tian2020road}. Besides that, driving simulators like CARLA \cite{Dosovitskiy17} and Airsim \cite{airsim2017fsr} are even more important over time as for scene replay, data augmentation and reinforcement learning \cite{chen2019lbc}.

\subsection{Relationship Prediction and Scene Graph Generation}

With the fast development of computer vision techniques, scene understanding --other than object detection-- has been the new frontier of computer vision. And it has been proved that such scene understanding tasks could also improve the performance of fundamental tasks like object detection\cite{johnson2015image}. From input/output aspect, scene understanding includes multiple tasks like image captioning, visual question answering (VQA)\cite{li2019relation, narasimhan2018out} and scene graph generation \cite{li2020learning, mylavarapu2020understanding, yu2020scene, zellers2018neural, li2017scene, xu2017scene}. In this paper, we are focusing on scene graph generation tasks. 

From the category aspect, there are common scene graph generation (relationship prediction) tasks and task-specific scene graph generation. Both of them benefit from the theoretical breakthrough of graph neural network and graph generation network\cite{simonovsky2018graphvae, simonovsky2017dynamic, li2015gated, bendimerad2019mining}.  In the common area, their models were trained by large open image datasets, such as Imagenet and MS COCO. Also, there are some dataset designed for this task, like Visual Genome \cite{krishna2017visual} and VRR-VG \cite{liang2019vrr}. However, for the task-specific aspect, such as road scene graph generation for intelligent vehicles, there are only several datasets and methods \cite{li2020learning, mylavarapu2020understanding, yu2020scene} that can fit the need.

\section{METHODS}
\label{s1:methods}

\subsection{Dataset setup}
\label{ss:datasetsetup}
Research in this paper an incremental work where we further exploit the potential of our dataset Road Scene Graph Dataset\cite{tian2020road}. Road Scene Graph Dataset includes 500 scenes, each one is around 20 seconds and sampled at 2\,Hz, so there are around 20,000 scene graphs in this dataset. The categories of objects and relationships are listed in table \ref{relationships}. Here, ``objects'' indicate the high-level categories of objects, e.g. ``Human'' includes pedestrians and workers, ``Obstacle'' includes traffic cones, debris, barriers, etc. For relationship data, we selected 21 kinds of common and basic relationships. 

\begin{table}[htbp]
    \caption{Objects and relationships in road scene graph dataset}
    \setlength{\tabcolsep}{5pt}
    \begin{tabular}{|c|c|c|c|c|}
    \hline
                                                                & Human & Vehicle                                                                                                                        & Obstacle                                                      & Traffic-sign                                                                      \\ \hline
    Human                                                       & Group & \begin{tabular}[c]{@{}c@{}}Behind\\ on-lane\\ waiting-for-cr\\ May-intersect\end{tabular}                                      & Behind                                                        & waiting-ts                                                                        \\ \hline
    \begin{tabular}[c]{@{}c@{}}Vehicle\\ (cyclist)\end{tabular} & --    & \begin{tabular}[c]{@{}c@{}}Group\\ Same-lane\\ Following\\ Approaching\\ Waiting-for-cr\\ Passing by\\ Overtaking\end{tabular} & \begin{tabular}[c]{@{}c@{}}passing-by\\ avoiding\end{tabular} & \begin{tabular}[c]{@{}c@{}}waiting-for-ts\\ stop-by-ts\\ react-by-ts\end{tabular} \\ \hline
    Obstacle                                                    & --    & --                                                                                                                             & Group                                                         & behind                                                                            \\ \hline
    Traffic-sign                                                & --    & --                                                                                                                             & --                                                            & --                                                                                \\ \hline
    \end{tabular}
    \centering
    \label{relationships}
\vspace{-1em}
\end{table}

\subsection{Road Scene graph Generation with Message Passing Network}

Road Scene Graph (RSG), which is a structured data representation of objects and their semantic relationships around the ego vehicle, can be defined as a tuple $G = <V, E>$.

\begin{itemize}
    \item  $V = \{v_1, v_2, \cdots, v_n\}$ stands for the object status. The nodes in $V$ are manually defined: $v_i = \langle{\mathbf{l}_{ni}, x_i, y_i, v_{xi}, v_{yi}, a_{xi}, a_{yi}, y, p, r}\rangle$, where $\mathbf{l_{ni}}$ represents the one-hot label of node's type, position ($x_i$, $y_i$),  velocity ($v_{xi}$, $v_{yi}$), acceleration ($a_{xi}$, $a_{yi}$), and heading angles (yaw $y$, pitch $p$, and roll $r$). In nuScenes dataset, we ignore the first and final frames as their object status are incomplete.
    \item  $E = \{ e_{ij} \; | \; v_i, v_j \in E  \}$ is the relationship set of the graph G. For the input graph $E_{in}$, the elements are $e_{i \to j} = (dx, dy, dvx, dvy, P(R|v_i, v_j) )$,  $dx, dy, dvx, dvy$ correspond to the position and velocity difference between two objects. For generated graph $E_{gen}$, the elements $ e_{i \to j} \in E_{gen} $ are the one-hot prediction vector indicates the predicted relationship.
\end{itemize}

For $ P(r_i \;| \; v_i, v_j) $, where $r_i \in R $, and $R$ is for the relationship set, it is well-known that the statistical information of object co-occurrence is a strong prior knowledge. So, inspired by \cite{chen2019knowledge}, we count the relationship occurrence in Road Scene Graph dataset as prior knowledge. For example, if $r_i$ means ``passing-by'' relationship and $v_i, v_j$ is for a human and a vehicle, then $P$ stand for the statistical co-occurrence of a vehicle passing by a pedestrian. In this way, we not only increase the performance but also significantly restrain the occurrence of incorrect relationships like ``$<barrier - passingby - another \; barrier>$''.

And for the graph generation network, our model was inspired by the graph gated network structure, a popular structure in multiple graph generation networks, to propagate node messages in graph \cite{xu2017scene, li2015gated, chen2019knowledge, khademi2020deep}.

\begin{figure*}[h]
    \centering
    \includegraphics[width=0.9\linewidth]{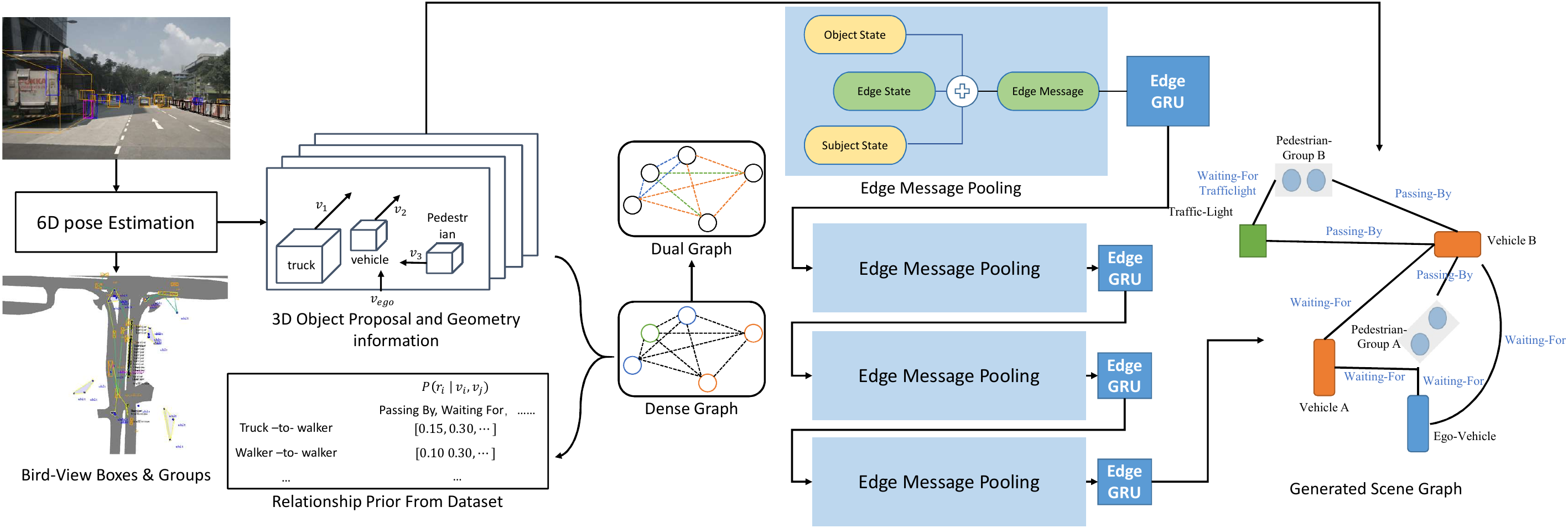}  
    \caption{ Representation of our graph prediction model. It takes object proposal as input and firstly generates a dense graph and its dual graph from these proposals. Then, we use a edge message pooling model and edge GRU to propagate messages through edges. Here we stacked the model 4 times. $\oplus$ symbol in edge message pooling model stands for a learnable parameter set. This model is inspired by \cite{xu2017scene}, which is an effective graph generation framework for common relationship prediction.}
    \label{fig:model}
\end{figure*}

Different from common scene graph generation problems, in this task we do not need to simultaneously predict object status and their relationships, as their bounding boxes could be obtained from perception module in a self-driving system, with LiDAR auxiliary bounding box regression, it would be very accurate. So, we focus instead on the edge prediction part. However, predicting edge is much more difficult than predicting node status, because compared to update only node status, update node and edge at the same time will bring additional complexity to the model and training process. \cite{xu2017scene} proposed an interesting insight that, if we take relationship as an individual node, then road scene graph will turn to a bipartite graph. In other words, node vector $V \in G$ could be separated into 2 sets: object set and relationship node set. So here we turn the original graph into its dual graph. And edge type prediction problem will be turned into a node prediction problem, which is much easier to solve. In \cite{xu2017scene}, the scene graph generation problem could be formulated as finding the optimal $x^{*} = \argmax_x Pr(x|I, B_I)$ given the image $I$ and bounding box proposals $B_I$. Therefore, $ Pr(x|I, B_I)$ can be defined as follows:

\begin{equation}
    Pr(x|I, B_I) = \prod_{i \in V} \prod_{j \neq i} Pr(x^{cls}_i, x^{bbox}_i, x_{i \to j} |I, B_I)
\end{equation}

In this paper, as we do not need bounding box detection. In contrast, we utilize the relationship prior distribution to improve accuracy. The road scene graph generation problem is:

\begin{equation}
    Pr(E_{gen}| V_I, P(r_i \;| \; v_i, v_j) ) = \prod_{i \in V} \prod_{j \neq i} Pr(e_{i \to j} |V_I, P(r_i \;| \; v_i, v_j) )
\end{equation}

Fig. \ref{fig:model}  illustrates our graph prediction network structure. This network takes the object proposal as input; the proposal includes bounding box, type, velocity, and acceleration. Another input is the prior probabilistic knowledge $P(r_i \;| \; v_i, v_j)$ form dataset. Then, the model generates a dense graph and its dual graph. This is a computational expensive operation, as for a graph with $n$ objects, the dense graph will contain $\frac{n(n-1)}{2}$ edges, and its dual graph will contain same amount of nodes. So, while we annotate data, we cluster the similar objects in groups to reduce graph size. As Fig. \ref{fig:model} bottom left illustrates, objects in yellow dash circle indicate that they belong to a group.

To enable message propagation on graphs, we choose edge GRU \cite{cho2014properties, li2015gated}  model, which is a simple but effective method. For each node in the dual graph, we set up a vector $h_t$ indicates its hidden status. As every node in dual graph (in other words, edge in scene graph) sharing the same update rule, we share the same set of parameters among all node GRUs. Eq. \ref{equ:gru} lists the update step formula for our GRU model:

\begin{equation}
    \begin{aligned}
        r_t &= \sigma(W_{xr} f_t + W_{hr} \hat{h}_{t-1}) \\
        z_t &= \sigma(W_{xz} f_t + W_{hz} \hat{h}_{t-1}) \\
        \widetilde{h}_t &= \tanh (W_{xh} f_t + W_{hh} (r_t \odot \hat{h}_{t-1} )) \\
        h_t &= (z_t \odot \hat{h}_{t-1} ) + (1 - z_t) \odot \widetilde{h}_t \\
        a_t &= \sigma(W_l h_t + b_l) \\
    \end{aligned}
    \label{equ:gru}
\end{equation}

Here, $\sigma()$ is for sigmoid function, and all $W$ is for learnable parameters. $\widetilde{h}_t$ are used as previous hidden state. Update gate $z_t$ has a role to adjust the previous information $\hat{h}_{t-1}$. Finally, binary output $a_t$ is obtained after a fully connected layer $W_l, b_l$.

\subsection{Metrics}

\begin{figure}[h]
    \centering
    \includegraphics[width=0.9\linewidth]{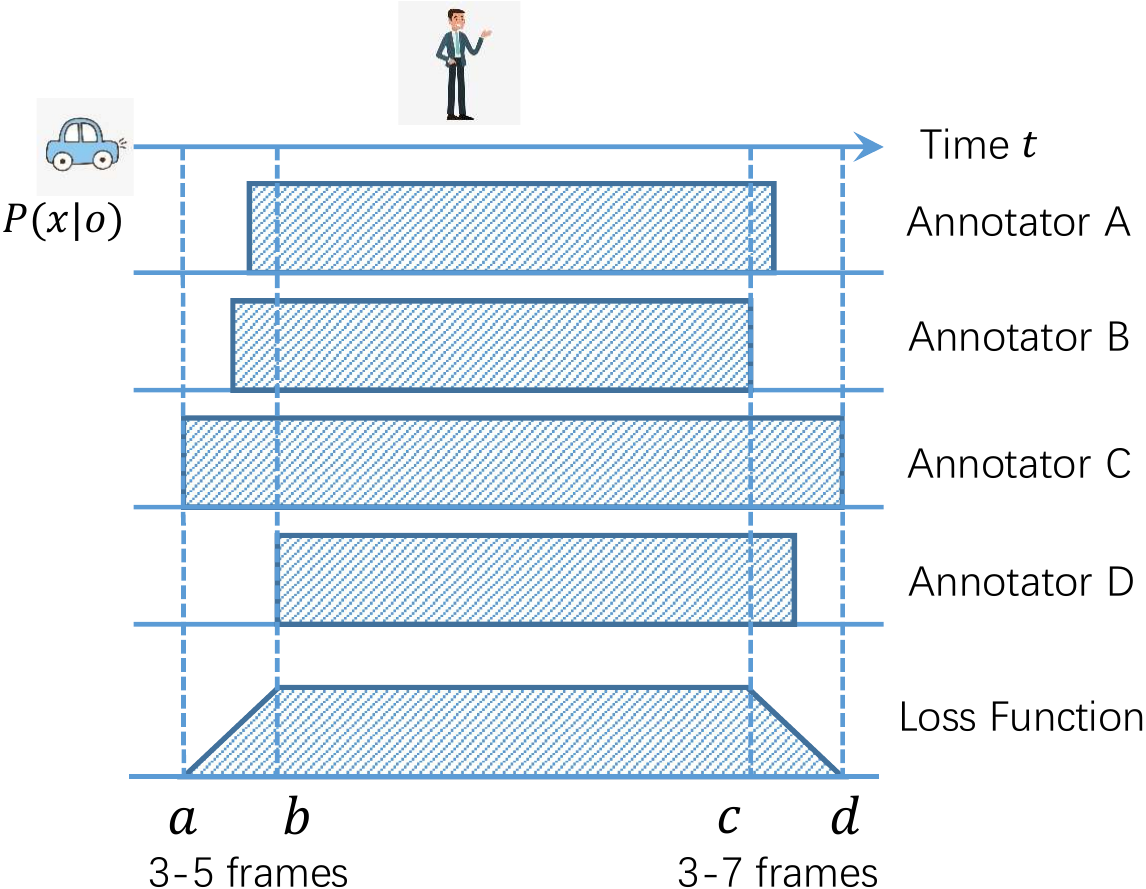}  
    \caption{ A brief example of annotation subjectivity in Road Scene Graph Dataset. Here the horizontal axis stands for the time (frames). While a car passing by a pedestrian, different annotators will propose slightly different annotations. }
    \label{fig:slope}
\end{figure}

Another fundamental difference between common scene graph prediction tasks and Road Scene Graph prediction is the time subjectivity. In other words, the starting time and ending time of a specific relationship may vary by different annotators. For example, as Fig. \ref{fig:slope} illustrates, when a vehicle passing by the pedestrian, starting and ending time among different annotator's annotations will vary. In common scene graph prediction tasks, this problem does not exist as it takes a single image, not a sequence of images as input. As a supplementary experiment, we invited 6 annotators, to label 10 relationships in 10 different scenes. The results show that the time gap will be about 3 to 7 frames (1.5 to 3.5 seconds) on average. To reduce the impact it makes into the model performance, we bring out the slope weight function. As Fig. \ref{fig:slope} illustrates, it brings a little bit relaxation at the start and end time of relationships. Eq. \ref{equ:slope} lists the slope weight function when ground truth is $\frac{b-a}{2}, \frac{d-c}{2}$. When calculating the loss function using sparse softmax cross entropy, we multiply this weight as a penalty item to every edge prediction. So the relationships at the ``slope'', whether successfully predicted or not, will influence the model less than normal relationships.

\begin{equation}
    \begin{aligned}
        w &= \frac{2}{d+c-a-b} \frac{x-a}{b-a} \quad &if \; x \in [a, b) \\
        w &= 1 \quad &if \; x \in [b, c) \\
        w &= \frac{2}{d+c-a-b} \frac{d-x}{d-c} \quad &if \; x \in [c, d] \\
    \end{aligned}
    \label{equ:slope}
\end{equation}  

We evaluate our method for generating scene graphs in two metrics: R@K metric\cite{lu2016visual} and Pairwise prediction accuracy. R@K metric is used to evaluate the difference of a generated graph to the ground truth, while pairwise prediction accuracy is used to evaluate the prediction accuracy for a specific kind of relationship, like ``waiting-for'' or ``passing-by''. Compared with normal metrics like mAP, R@K metric is significantly better. As in most of the time the scene graph is a sparse graph, more than 95\% percent of relationships are actually ``No relation''. Metrics like mAP would falsely penalize positive predictions on unlabeled relationships, and suppress the form of relationship.


\section{EXPERIMENT}
\label{s:experiment}
\begin{figure}[h]
    \centering
    \includegraphics[width=0.9\linewidth]{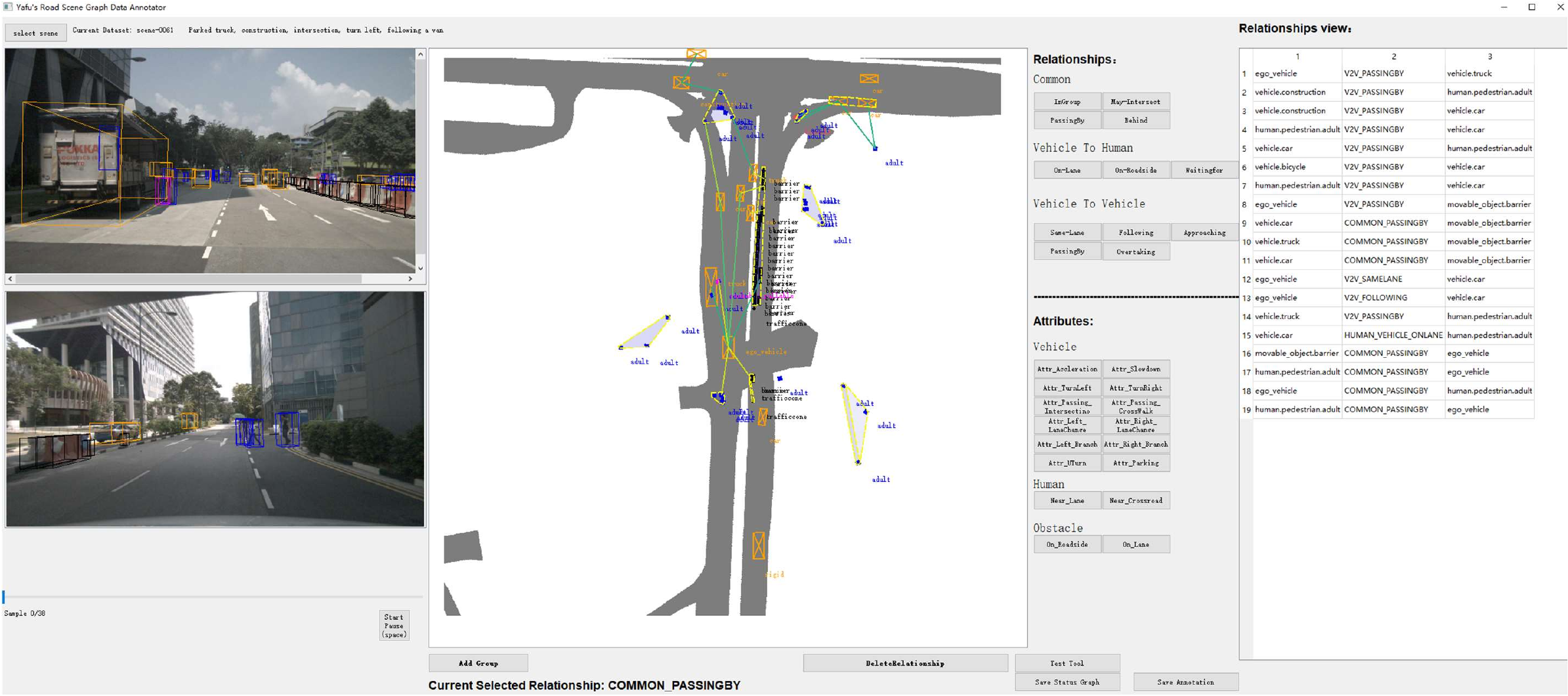}  
    \caption{ Road Scene Graph annotator we used to set up experiment dataset. This annotator was created to enable annotating relationships and group information in nuScenes dataset.}
    \label{fig:annotator}
\end{figure}

Similar to our prior work \cite{tian2020road}, we generated Road Scene Graph dataset for training and evaluation, this dataset has been introduced in Section~\ref{ss:datasetsetup}. Fig. \ref{fig:annotator} illustrates the user interface of our data annotator. We set the final 10\% scenes in dataset as test set, and use the remaining for training set.

\begin{figure}[htbp]
    \centering
    \includegraphics[width=1\linewidth]{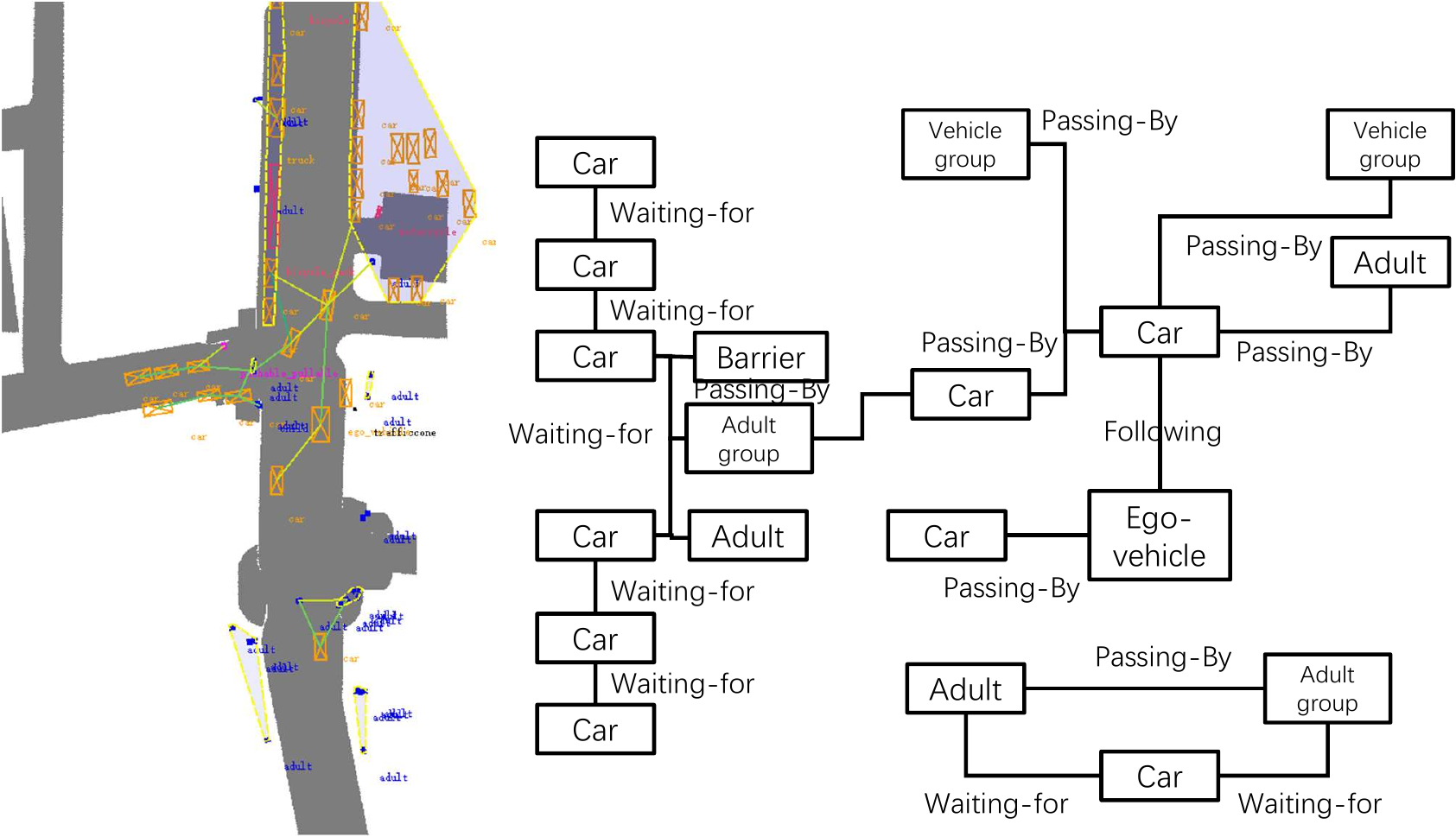}  
    \caption{(Left)~Bird eye view of the ego-vehicle and object nearby, yellow dash circle indicates group border, and colored lines stand for the relationships. (Right)~Generated scene graph sample. In order to keep spatial position and make graph clear to read, we manually draw it here.}
    \label{fig:sample}
\end{figure}

Fig. \ref{fig:sample} illustrates a sample frame in nuScenes, and its corresponding road scene graph (scene number 0103, frame 21/39).  Here, the ego vehicle is following the vehicle in front of it, and the vehicle on the opposite lane is waiting for pedestrian crossing the road while trying to turn right. Vehicles on the left branch are also waiting for the pedestrians. By arranging all objects into a graph, we can visualize a very complex traffic scene and objects' interaction. Also, this graph data representation could be easily processed by graph neural networks. 

Compared with our previous methods \cite{tian2020road}, our current model performs significantly better on R@K recall and Pairwise accuracy. Table \ref{tab:01} lists the R@K metric and Pairwise accuracy performance of our model. Here the NGPGV (Next graph prediction using graph VGA) and NGPGVEB (Next graph prediction using graph VGAE and bird's-eye view ) come from our previous work. And they take a status graph $G'$ as input and predict the graph of the next timestep.

In our Road Scene Graph Dataset, the average of edge amount is 19 per frame, the average degree of scene graph dataset is 2.43 and our predicted scene graph is 1.85. This means that, although we use @K metric to evaluate the graph, the graph generation process is still conservative, and generates fewer relationships than ground truth.

\begin{table}[htbp]
    \caption{Results of relationship prediction approach, compared with our previous work \cite{tian2020road}.}
    \begin{tabular}{rcccc}
    \cline{1-4}
    Methods         & R@15 & R@25 & Avg. Pairwise Accuracy &  \\ \cline{1-4}
    NGPGV  \cite{tian2020road}         & 26.8 & 34.8 & 37.0                      &  \\
    NGVPGEB \cite{tian2020road}       & 13.5 & 35.7 & 26.5                      &  \\
    RSG-Net         & 31.4 & 44.4 & 55.0                      &  \\
    RSG-Net (Slope) & \textbf{41.6} & \textbf{47.1} & \textbf{59.5}                      &  \\ \cline{1-4}
    \end{tabular}
    \label{tab:01}
\end{table}

\begin{figure}[htbp]
    \centering
    \includegraphics[width=1\linewidth]{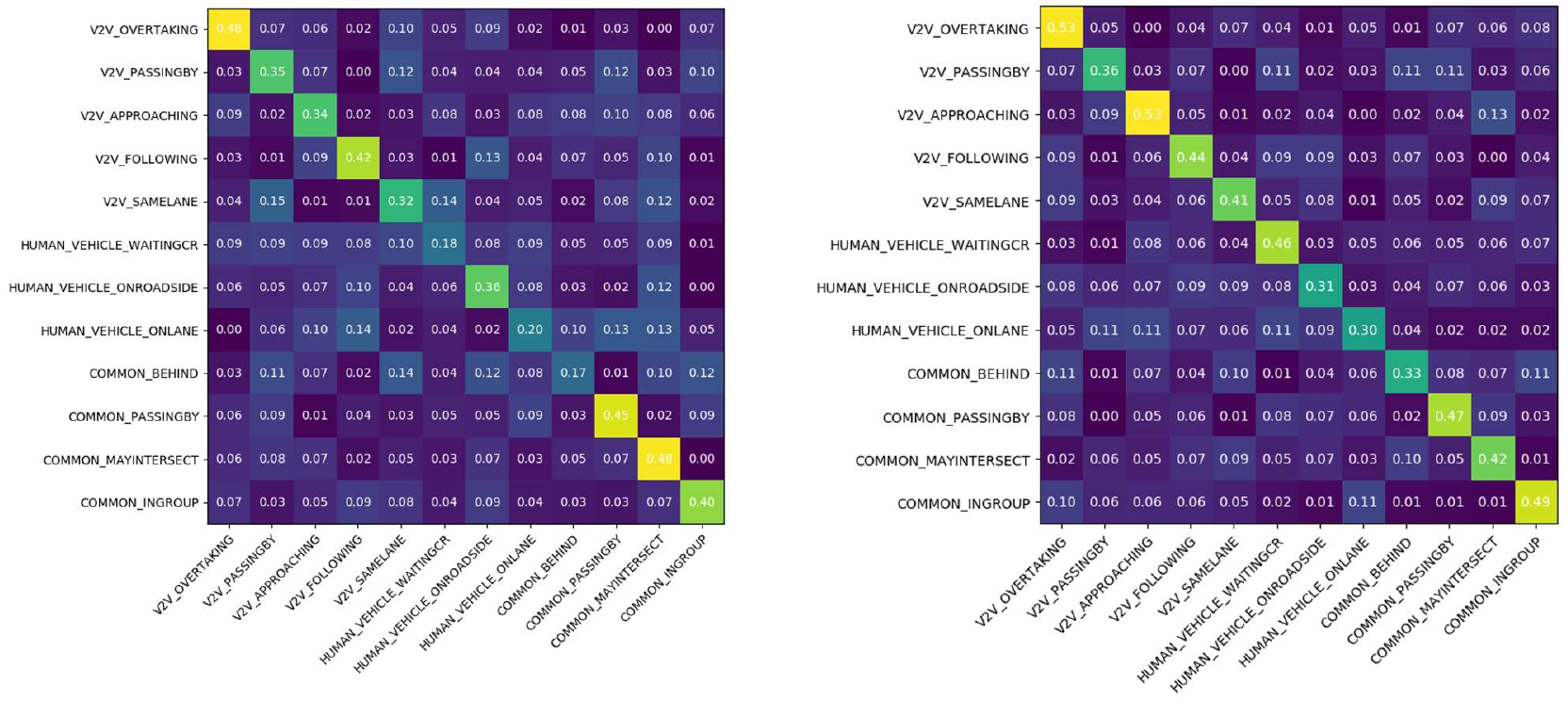}  
    \caption{Matrix of the error prediction distribution among the relationships:  (Left) using normal loss penalty, (Right) using slope loss penalty. Some relationships have been combined for simplicity, e.g., ``human-behind-vehicle'', ``vehicle-behind-vehicle'', and ``vehicle-behind-barrier'' were combined as ``common-behind''.}
    \label{fig:accMatrix}
\end{figure}

Fig. \ref{fig:accMatrix} illustrates the pairwise accuracy and wrong prediction distribution of our model. The main diagonal element is for the pairwise accuracy of correct relationship prediction, and other elements stand for corresponding wrong predictions, e.g., the relationship ``overtaking'' classified as approaching (the error rate in column 3 row 1). By using slope penalty, the Pairwise accuracy increased about 4.5\%. However, the imbalanced data class is still a problem: as in Road Scene Graph Dataset the top 5 relationships take 84\% occurrence. The model tends to predict very common relationship categories other than very rare ones (ex., ``overtaking'', since scene duration is 20\,seconds, it is quite rare in nuScenes dataset). 

\begin{figure}[htbp]
    \centering
    \includegraphics[width=0.7\linewidth]{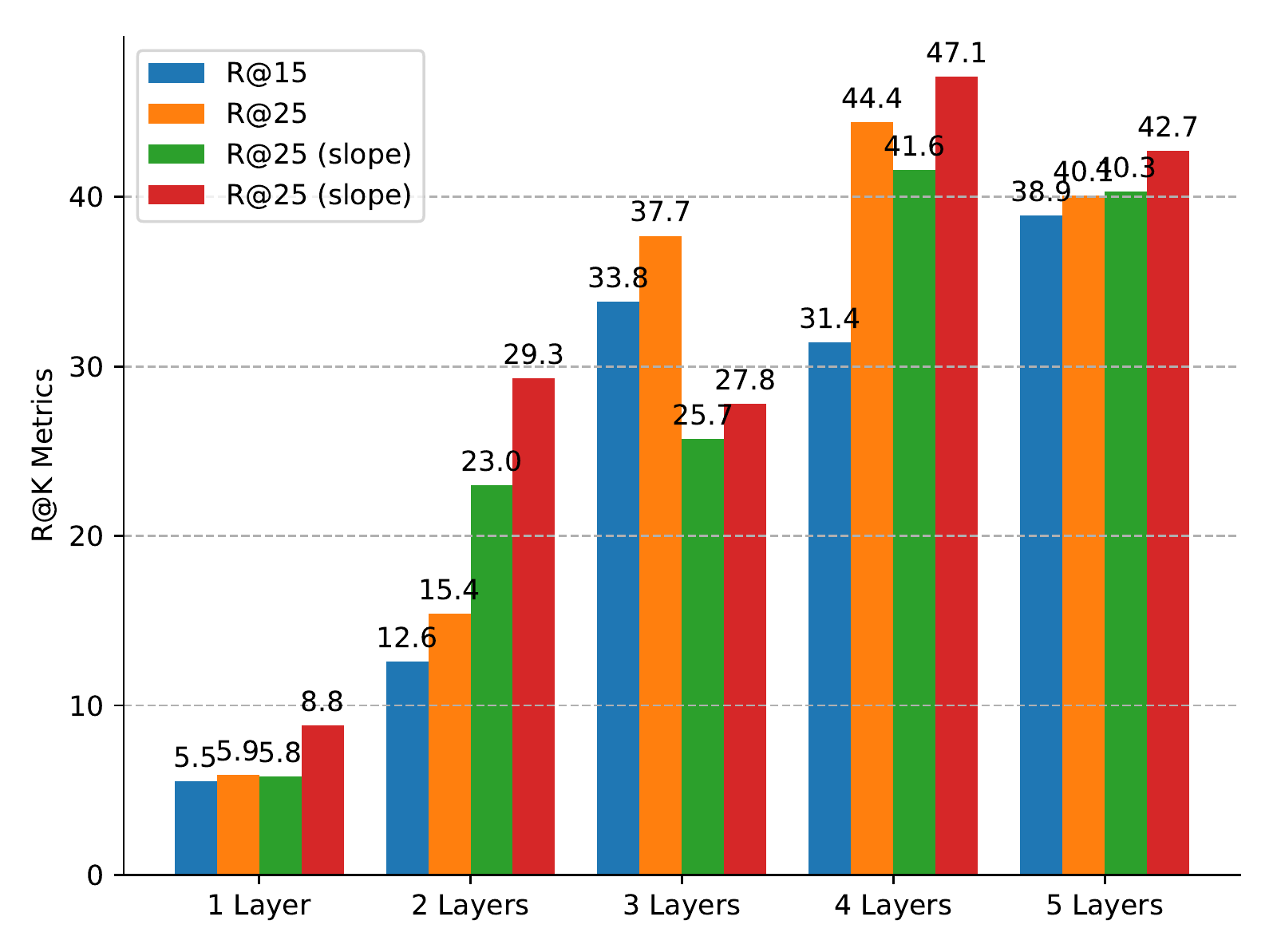}  
    \caption{R@K metrics performance using our models with different layers of network. As Fig. \ref{fig:model} illustrates, each layer includes a pooling and an edge GRU module. The peak performance appears with slope penalty while we stacked 4 layers GRU module. } 
    \label{fig:layerRK}
\end{figure}

Regarding the network structure, Fig. \ref{fig:layerRK} illustrates the R@K metrics by changing different depth networks, with or without slope penalty. As Road Scene Graph Dataset \cite{tian2020road} is still a small dataset, we only stack very few layers to get optimum performance. 

\section{CONCLUSIONS}
\label{s:conclusions}
In this paper, we proposed RSG-Net, s a graph convolutional network to model fundamental semantic relationships between vehicles, pedestrians, and other objects, and arrange them into a graph structure. Experiments conducted on Road Scene Graph Dataset indicate that our model could capture and predict semantic relational data. In future works, this road scene graph could be helpful for multiple tasks like potential risk detection, road scene captioning, etc. Besides that, the performance of our current model could be further optimized with the rapid theoretical breakthrough of graph networks.

 



\section*{ACKNOWLEDGMENT}

This work was supported by the Science Fund for Creative
Research Groups of the National Natural Science Foundation
of China under grant number 51521003. The research work
carried out at the State Key Laboratory of Robotic and
Intelligent System, Harbin Institute of Technology. We are
deeply grateful for the kind cooperation of both faculty and
students.


\bibliography{ref}

\begin{thebibliography}{10}
\providecommand{\url}[1]{#1}
\csname url@samestyle\endcsname
\providecommand{\newblock}{\relax}
\providecommand{\bibinfo}[2]{#2}
\providecommand{\BIBentrySTDinterwordspacing}{\spaceskip=0pt\relax}
\providecommand{\BIBentryALTinterwordstretchfactor}{4}
\providecommand{\BIBentryALTinterwordspacing}{\spaceskip=\fontdimen2\font plus
\BIBentryALTinterwordstretchfactor\fontdimen3\font minus
  \fontdimen4\font\relax}
\providecommand{\BIBforeignlanguage}[2]{{%
\expandafter\ifx\csname l@#1\endcsname\relax
\typeout{** WARNING: IEEEtran.bst: No hyphenation pattern has been}%
\typeout{** loaded for the language `#1'. Using the pattern for}%
\typeout{** the default language instead.}%
\else
\language=\csname l@#1\endcsname
\fi
#2}}
\providecommand{\BIBdecl}{\relax}
\BIBdecl

\bibitem{ramanishka2018CVPR}
V.~Ramanishka, Y.-T. Chen, T.~Misu, and K.~Saenko, ``Toward driving scene
  understanding: A dataset for learning driver behavior and causal reasoning,''
  in \emph{Conference on Computer Vision and Pattern Recognition (CVPR)}, 2018.

\bibitem{tian2020road}
Y.~Tian, A.~Carballo, R.~Li, and K.~Takeda, ``Road scene graph: A semantic
  graph-based scene representation dataset for intelligent vehicles,''
  \emph{arXiv preprint arXiv:2011.13588}, 2020.

\bibitem{caesar2020nuscenes}
H.~Caesar, V.~Bankiti, A.~H. Lang, S.~Vora, V.~E. Liong, Q.~Xu, A.~Krishnan,
  Y.~Pan, G.~Baldan, and O.~Beijbom, ``nuscenes: A multimodal dataset for
  autonomous driving,'' in \emph{Proceedings of the IEEE/CVF Conference on
  Computer Vision and Pattern Recognition}, 2020, pp. 11\,621--11\,631.

\bibitem{kuo2015deepbox}
W.~Kuo, B.~Hariharan, and J.~Malik, ``Deepbox: Learning objectness with
  convolutional networks,'' in \emph{Proceedings of the IEEE international
  conference on computer vision}, 2015, pp. 2479--2487.

\bibitem{cho2014multi}
H.~Cho, Y.-W. Seo, B.~V. Kumar, and R.~R. Rajkumar, ``A multi-sensor fusion
  system for moving object detection and tracking in urban driving
  environments,'' in \emph{2014 IEEE International Conference on Robotics and
  Automation (ICRA)}.\hskip 1em plus 0.5em minus 0.4em\relax IEEE, 2014, pp.
  1836--1843.

\bibitem{xu2018pointfusion}
D.~Xu, D.~Anguelov, and A.~Jain, ``Pointfusion: Deep sensor fusion for 3d
  bounding box estimation,'' in \emph{Proceedings of the IEEE Conference on
  Computer Vision and Pattern Recognition}, 2018, pp. 244--253.

\bibitem{liang2020learning}
M.~Liang, B.~Yang, R.~Hu, Y.~Chen, R.~Liao, S.~Feng, and R.~Urtasun, ``Learning
  lane graph representations for motion forecasting,'' in \emph{European
  Conference on Computer Vision}.\hskip 1em plus 0.5em minus 0.4em\relax
  Springer, 2020, pp. 541--556.

\bibitem{kim2017interpretable}
J.~Kim and J.~Canny, ``Interpretable learning for self-driving cars by
  visualizing causal attention,'' in \emph{Proceedings of the IEEE
  international conference on computer vision}, 2017, pp. 2942--2950.

\bibitem{kim2019grounding}
J.~Kim, T.~Misu, Y.-T. Chen, A.~Tawari, and J.~Canny, ``Grounding
  human-to-vehicle advice for self-driving vehicles,'' in \emph{Proceedings of
  the IEEE/CVF Conference on Computer Vision and Pattern Recognition}, 2019,
  pp. 10\,591--10\,599.

\bibitem{kim2018textual}
J.~Kim, A.~Rohrbach, T.~Darrell, J.~Canny, and Z.~Akata, ``Textual explanations
  for self-driving vehicles,'' in \emph{Proceedings of the European conference
  on computer vision (ECCV)}, 2018, pp. 563--578.

\bibitem{li2020learning}
C.~Li, Y.~Meng, S.~H. Chan, and Y.-T. Chen, ``Learning 3d-aware egocentric
  spatial-temporal interaction via graph convolutional networks,'' in
  \emph{2020 IEEE International Conference on Robotics and Automation
  (ICRA)}.\hskip 1em plus 0.5em minus 0.4em\relax IEEE, 2020, pp. 8418--8424.

\bibitem{mylavarapu2020understanding}
S.~Mylavarapu, M.~Sandhu, P.~Vijayan, K.~M. Krishna, B.~Ravindran, and
  A.~Namboodiri, ``Understanding dynamic scenes using graph convolution
  networks,'' \emph{arXiv preprint arXiv:2005.04437}, 2020.

\bibitem{yu2020scene}
S.-Y. Yu, A.~V. Malawade, D.~Muthirayan, P.~P. Khargonekar, and M.~A.~A.
  Faruque, ``Scene-graph augmented data-driven risk assessment of autonomous
  vehicle decisions,'' \emph{arXiv preprint arXiv:2009.06435}, 2020.

\bibitem{behley2019semantickitti}
J.~Behley, M.~Garbade, A.~Milioto, J.~Quenzel, S.~Behnke, C.~Stachniss, and
  J.~Gall, ``Semantickitti: A dataset for semantic scene understanding of lidar
  sequences,'' in \emph{Proceedings of the IEEE International Conference on
  Computer Vision}, 2019, pp. 9297--9307.

\bibitem{sun2020scalability}
P.~Sun, H.~Kretzschmar, X.~Dotiwalla, A.~Chouard, V.~Patnaik, P.~Tsui, J.~Guo,
  Y.~Zhou, Y.~Chai, B.~Caine \emph{et~al.}, ``Scalability in perception for
  autonomous driving: Waymo open dataset,'' in \emph{Proceedings of the
  IEEE/CVF Conference on Computer Vision and Pattern Recognition}, 2020, pp.
  2446--2454.

\bibitem{wolfe2020rapid}
B.~Wolfe, B.~Seppelt, B.~Mehler, B.~Reimer, and R.~Rosenholtz, ``Rapid holistic
  perception and evasion of road hazards.'' \emph{Journal of experimental
  psychology: general}, vol. 149, no.~3, p. 490, 2020.

\bibitem{Dosovitskiy17}
A.~Dosovitskiy, G.~Ros, F.~Codevilla, A.~Lopez, and V.~Koltun, ``{CARLA}: {An}
  open urban driving simulator,'' in \emph{Proceedings of the 1st Annual
  Conference on Robot Learning}, 2017, pp. 1--16.

\bibitem{airsim2017fsr}
\BIBentryALTinterwordspacing
S.~Shah, D.~Dey, C.~Lovett, and A.~Kapoor, ``Airsim: High-fidelity visual and
  physical simulation for autonomous vehicles,'' in \emph{Field and Service
  Robotics}, 2017. [Online]. Available: \url{https://arxiv.org/abs/1705.05065}
\BIBentrySTDinterwordspacing

\bibitem{chen2019lbc}
D.~Chen, B.~Zhou, V.~Koltun, and P.~Kr\"ahenb\"uhl, ``Learning by cheating,''
  in \emph{Conference on Robot Learning (CoRL)}, 2019.

\bibitem{johnson2015image}
J.~Johnson, R.~Krishna, M.~Stark, L.-J. Li, D.~Shamma, M.~Bernstein, and
  L.~Fei-Fei, ``Image retrieval using scene graphs,'' in \emph{Proceedings of
  the IEEE conference on computer vision and pattern recognition}, 2015, pp.
  3668--3678.

\bibitem{li2019relation}
L.~Li, Z.~Gan, Y.~Cheng, and J.~Liu, ``Relation-aware graph attention network
  for visual question answering,'' in \emph{Proceedings of the IEEE
  International Conference on Computer Vision}, 2019, pp. 10\,313--10\,322.

\bibitem{narasimhan2018out}
M.~Narasimhan, S.~Lazebnik, and A.~Schwing, ``Out of the box: Reasoning with
  graph convolution nets for factual visual question answering,'' in
  \emph{Advances in neural information processing systems}, 2018, pp.
  2654--2665.

\bibitem{zellers2018neural}
R.~Zellers, M.~Yatskar, S.~Thomson, and Y.~Choi, ``Neural motifs: Scene graph
  parsing with global context,'' in \emph{Proceedings of the IEEE Conference on
  Computer Vision and Pattern Recognition}, 2018, pp. 5831--5840.

\bibitem{li2017scene}
Y.~Li, W.~Ouyang, B.~Zhou, K.~Wang, and X.~Wang, ``Scene graph generation from
  objects, phrases and region captions,'' in \emph{Proceedings of the IEEE
  International Conference on Computer Vision}, 2017, pp. 1261--1270.

\bibitem{xu2017scene}
D.~Xu, Y.~Zhu, C.~B. Choy, and L.~Fei-Fei, ``Scene graph generation by
  iterative message passing,'' in \emph{Proceedings of the IEEE conference on
  computer vision and pattern recognition}, 2017, pp. 5410--5419.

\bibitem{simonovsky2018graphvae}
M.~Simonovsky and N.~Komodakis, ``Graphvae: Towards generation of small graphs
  using variational autoencoders,'' in \emph{International Conference on
  Artificial Neural Networks}.\hskip 1em plus 0.5em minus 0.4em\relax Springer,
  2018, pp. 412--422.

\bibitem{simonovsky2017dynamic}
------, ``Dynamic edge-conditioned filters in convolutional neural networks on
  graphs,'' in \emph{Proceedings of the IEEE conference on computer vision and
  pattern recognition}, 2017, pp. 3693--3702.

\bibitem{li2015gated}
Y.~Li, D.~Tarlow, M.~Brockschmidt, and R.~Zemel, ``Gated graph sequence neural
  networks,'' \emph{arXiv preprint arXiv:1511.05493}, 2015.

\bibitem{bendimerad2019mining}
A.~Bendimerad, ``Mining useful patterns in attributed graphs,'' Ph.D.
  dissertation, Universit{\'e} de Lyon, 2019.

\bibitem{krishna2017visual}
R.~Krishna, Y.~Zhu, O.~Groth, J.~Johnson, K.~Hata, J.~Kravitz, S.~Chen,
  Y.~Kalantidis, L.-J. Li, D.~A. Shamma \emph{et~al.}, ``Visual genome:
  Connecting language and vision using crowdsourced dense image annotations,''
  \emph{International journal of computer vision}, vol. 123, no.~1, pp. 32--73,
  2017.

\bibitem{liang2019vrr}
Y.~Liang, Y.~Bai, W.~Zhang, X.~Qian, L.~Zhu, and T.~Mei, ``Vrr-vg: Refocusing
  visually-relevant relationships,'' in \emph{Proceedings of the IEEE
  International Conference on Computer Vision}, 2019, pp. 10\,403--10\,412.

\bibitem{chen2019knowledge}
T.~Chen, W.~Yu, R.~Chen, and L.~Lin, ``Knowledge-embedded routing network for
  scene graph generation,'' in \emph{Proceedings of the IEEE/CVF Conference on
  Computer Vision and Pattern Recognition}, 2019, pp. 6163--6171.

\bibitem{khademi2020deep}
M.~Khademi and O.~Schulte, ``Deep generative probabilistic graph neural
  networks for scene graph generation,'' in \emph{Proceedings of the AAAI
  Conference on Artificial Intelligence}, vol.~34, no.~07, 2020, pp.
  11\,237--11\,245.

\bibitem{cho2014properties}
K.~Cho, B.~Van~Merri{\"e}nboer, D.~Bahdanau, and Y.~Bengio, ``On the properties
  of neural machine translation: Encoder-decoder approaches,'' \emph{arXiv
  preprint arXiv:1409.1259}, 2014.

\bibitem{lu2016visual}
C.~Lu, R.~Krishna, M.~Bernstein, and L.~Fei-Fei, ``Visual relationship
  detection with language priors,'' in \emph{European conference on computer
  vision}.\hskip 1em plus 0.5em minus 0.4em\relax Springer, 2016, pp. 852--869.

\end{thebibliography}

\end{document}